\documentclass[runningheads]{llncs}
\usepackage[T1]{fontenc}

\usepackage{graphicx}
\usepackage{rotating}
\usepackage{hyperref}

\usepackage[ruled,linesnumbered,lined]{algorithm2e}
\usepackage{amsmath}
\usepackage{tabularx}

\usepackage[compatibility=false]{caption}
\usepackage{subcaption}

\usepackage{graphicx}
\begin{document}
\title{TOPSIS-like metaheuristic for LABS problem}
%
%
\author{Aleksandra Urbańczyk\inst{1} \orcidID{0000-0002-6040-554X} \and
Bogumiła Papiernik\inst{1} \and
Piotr Magiera\inst{1}
\and Piotr Urbańczyk\inst{2,1}\orcidID{0000-0001-8838-2354}
\and Aleksander Byrski\inst{1}\orcidID{0000-0001-6317-7012}}
\authorrunning{A. Urbańczyk et al.}
%
\institute{AGH University of Krakow, Al. Mickiewicza 30, 30-059 Krakow, Poland
\email{\{aurbanczyk,purbanczyk,olekb\}@agh.edu.pl}, \email{\{bpapiernik,magiera\}@student.agh.edu.pl} \and
Jagiellonian University, ul. Gołębia 24, 31-007 Kraków, Poland
\email{piotr.urbanczyk@uj.edu.pl}}
%
\maketitle              
\begin{abstract}

This paper presents the application of socio-cognitive mutation operators inspired by the TOPSIS method to the Low Autocorrelation Binary Sequence (LABS) problem. Traditional evolutionary algorithms, while effective, often suffer from premature convergence and poor exploration-exploitation balance. To address these challenges, we introduce socio-cognitive mutation mechanisms that integrate strategies of following the best solutions and avoiding the worst. By guiding search agents to imitate high-performing solutions and avoid poor ones, these operators enhance both solution diversity and convergence efficiency. Experimental results demonstrate that TOPSIS-inspired mutation outperforms the base algorithm in optimizing LABS sequences. The study highlights the potential of socio-cognitive learning principles in evolutionary computation and suggests directions for further refinement.

\end{abstract}

\section{Introduction}
Optimization problems play a fundamental role in various scientific and engineering fields, with applications ranging from telecommunications to artificial intelligence. One particularly challenging problem is the Low-Autocorrelation Binary Sequence (LABS) problem, which requires finding binary sequences that minimize autocorrelation \cite{golay}. LABS is known for its highly rugged search space, making it difficult for traditional optimization methods to find optimal or near-optimal solutions efficiently.

To address the challenges posed by LABS, researchers have explored various metaheuristic algorithms  \cite{brest2018heuristic}, \cite{bovskovic2017low}, \cite{brest2020computational}. These methods, despite being quite successful in solving hard computational problems, are also known to struggle with premature convergence, exploration-exploitation imbalance, and difficulty escaping local optima. Among researchers experimenting with classic and new metaheuristics, there is always hope that new ideas drawn from natural phenomena can bring even a minor but repeatable improvement in the performance of optimization algorithms. Because of that, they tweak and change existing metaheuristics, they hybridize them, and they design new creative ones \cite{Biełaszek02012023}. By this means, they make algorithms that are better suited for solved problems than algorithms used before, and by that, the science progresses.
One group of such successful inspirations that improve existing metaheuristics are modifications based on the influence from socio-cognitive phenomena \cite{book}. Such phenomena provide valuable inspiration for evolutionary metaheuristics, as they highlight key mechanisms such as knowledge sharing, adaptive decision making, and social influence that can be incorporated into optimization algorithms. One approach to achieving this integration is through the development of novel mutation operators.

\section{State of the Art}

There are well-known metaheuristics that aim to redefine mutation beyond simple random bit flips or Gaussian noise.
An example of such modification consists of adaptive mutation strategies \cite{Blum2012AdaptiveMS}. Instead of using a fixed mutation rate, these approaches dynamically adjust mutation probabilities based on population diversity or fitness trends. For instance, in Evolution Strategies (ES), each individual can carry not only the object variables (i.e., the solution encoding) but also the strategy parameters. These parameters are subject to mutation and recombination along with the object variables, allowing the algorithm to adapt its search behavior during the optimization process \cite{MICHALEWICZ2003259}. Another example is Covariance Matrix Adaptation - Evolution Strategy (CMA-ES), that adapts the mutation process by sampling from a learned covariance matrix that describes the distribution of successful past mutations \cite{cma-es}.

Despite such enhancements, traditional evolutionary algorithms lack explicit mechanisms for knowledge sharing, imitation, and social adaptation, motivating the development of socio-cognitive metaheuristics.

\subsection{Socio-cognitively inspired hybrid mutation operators}
Socio-cognitive metaheuristics are an emerging class of optimization algorithms inspired by social learning, cognitive processes, and collective intelligence. Unlike classical metaheuristics that rely solely on stochastic operators (mutation, crossover, selection), socio-cognitive approaches incorporate adaptive learning mechanisms influenced by how humans and social groups acquire and share knowledge. 

They all relate to Albert Bandura’s Social Learning Theory \cite{bandura1986}, which posits that individuals acquire knowledge not only through direct experience but also by observing others, imitating behaviors, and interacting within a social environment. Bandura’s work introduced the idea that learning is an active, cognitive process influenced by attention, retention, reproduction, and motivation.  A key element of this theory is the concept of observational learning (modeling), which suggests that individuals learn new behaviors by watching the actions and consequences experienced by others.

In the context of computational intelligence and evolutionary metaheuristics, Bandura’s principles inspire socio-cognitive algorithms, where individuals (solutions) adapt by learning from successful peers, avoiding ineffective strategies, and dynamically adjusting search behaviors. This has been applied in various hybrid optimization techniques, such as socio-cognitive Ant Colony Optimization \cite{scaco} and imitation-based mutation operators, described below.

There are three closely related research papers that incorporate ideas similar to the proposed ones, but apply them to different basic metaheuristics and to different optimization problems. 

All of them, beside being inspired by socio-cognitive phenomena bear a resemblance to the TOPSIS (Technique for Order of Preference by Similarity to Ideal Solution) \cite{topsis81} method, widely used multi-criteria decision-making (MCDM) approach. This method based on the concept that the best solution should have the shortest distance from an ideal (optimal) solution and the greatest distance from the worst (non-optimal) solution. When incorporated into optimization algorithms, it can guide search toward promising regions of the solution space by encouraging movement toward high-quality solutions while avoiding poor ones.

The first research paper explores the socio-cognitive hybridization of the standard versions of Evolution Strategies (ES) by introducing learning-inspired mutation operators, which allow information exchange between individuals \cite{urbanczyk2021socio}.  
Three variations of socio-cognitive mutation are introduced:  
\begin{itemize}
    \item Closer-to-best: Individuals move toward historical top-performing solutions.
    \item Farther-from-worst: Solutions are modified to avoid poor solutions, reducing premature convergence.
    \item Both features: A combination of both strategies improves search adaptability.
\end{itemize}
These mechanisms were tested on standard single-objective benchmark optimization problems. The variant of mutation based on the imitation of the best individuals provided the best results, but combining it with the strategy of avoiding the worst solution was also promising. 

The second example is socio-cognitive optimization of Time-Delay Control Problems \cite{caste}. This study proposes a socio-cognitive evolutionary metaheuristic with a caste-based learning structure, where individuals belong to groups that interact and exchange genetic information.  
The mutation operator is adjusted to allow knowledge transfer between castes, enabling:  
\begin{itemize}
    \item Selective imitation of elite individuals.
    \item Avoidance of low-performing solutions through a dynamic repulsion mechanism.
\end{itemize}
It also introduces a TOPSIS-based mutation mechanism, in which individuals have a chance to be pulled towards the best solution/point and pushed away from the worst solution/point. 
The algorithm inspired by caste societies yielded superior results, while the TOPSIS-based approach also outperformed the basic Evolutionary Algorithm. In the discussion section, it was recommended for further investigation. 

The third work is the most sophisticated, and it takes the $(\mu+\lambda)$ Evolution Strategy as a basis for the implementation of several variants of socio-cognitive mutation \cite{ES_SC_article} and applies them to popular benchmark one-criterion optimization functions.  
It introduces, similarly to the works already mentioned, mechanisms working in two directions (mimicking the best and avoiding the worst solutions), but does it in a comprehensive and exhaustive way. Each direction is tested with two alternative methods, also considering possible combinations. In the most successful strategy for following the best individuals, genes are selectively modified based on their importance in the top solutions, determined by standard deviation weighting. The most successful repelling strategy moves solutions away from poor solutions using an inverse-square law repulsion mechanism.
The mechanisms presented in this paper were adapted for binary optimization problems and tested on the LABS problem. The following sections provide detailed methodology and results.

\section{TOPSIS-like mutation in evolutionary algorithm}
\subsection{Baseline evolutionary algorithm}
Genetic Algorithms (GAs) constitute a prominent class of evolutionary optimization techniques grounded in biological principles of natural selection \cite{holland1975adaptation,goldberg1989genetic}. In these algorithms, solution candidates are encoded as chromosome-like data structures, and a multiset of such solutions called population is iteratively evolved via the application of stochastic genetic operators. The evolution of a population $P$ to a new population $P'$ can be represented as:
\begin{equation}
    P' = \mu\left(s\left(f(P)\right)\right), 
\end{equation}
where $f$ is a fitness
function that evaluates the population members and returns a vector of values that measure the quality of each solution, $\mu$ is a composite function that introduces random variations to a subset of individuals within the population, usually through crossover and mutation operators, and $s$ is a selection function.
\enlargethispage{1.5\baselineskip}

\begin{algorithm}[h]
\caption{Genetic Algorithm (GA)}\label{algo:ga}
\KwIn{Population size $N$, crossover rate $p_c$, mutation rate $p_m$}
\KwOut{Best solution found}
Initialize population $P$ with $N$ individuals\;
Evaluate fitness of each individual in $P$\;
\While{Termination criterion is not met}{
    Select parents from population $P$ based on their fitness\;
    Apply crossover to selected parents with probability $p_c$ to produce offspring\;
    Apply mutation to offspring with probability $p_m$\;
    \tcp{Apply eventual TOPSIS-inspired variation operators here}
    Evaluate fitness of offspring\;
    Replace less fit individuals in $P$ with offspring to create new population\;
}
\Return the best individual from the final population\;
\end{algorithm}

The overall structure of the Genetic Algorithm (GA) is outlined in the algorithm \ref{algo:ga} schema.
This work presents a specialized GA implementation tailored for optimization in a binary search space, namely for solving the low autocorrelation binary sequence (LABS) problem, employing a deterministic selection strategy combined with classical variation operators.

\paragraph{Best Solution Selection}
The implemented selection mechanism is an elitist strategy through the best solution selection operator:
\begin{equation}
s\left(f(P)\right) = \{ x \in P \mid \forall y \in P, \, f(x) \geq f(y) \}.
\end{equation}
This means that the selection operator $s$ chooses the individual $x$ from the population $P$ whose fitness is greater than or equal to that of all other individuals in $P$.

\paragraph{Single-Point Crossover}
The implemented crossover method is a common single-point crossover that combines two parent solutions to produce offspring by exchanging segments of their genetic material at a crossover point chosen uniformly at random. For binary-encoded parents $p_1$ and $p_2$ of length $l$ 
and randomly selected crossover point 
$c \sim \mathcal{U}\{1, l-1\}$:
\begin{equation}
\begin{aligned}
o_1 &= \bigl(p_1[1:c],\, p_2[c+1:l]\bigr), \\
o_2 &= \bigl(p_2[1:c],\, p_1[c+1:l]\bigr).
\end{aligned}
\end{equation}

\paragraph{Bit-Flip Mutation}
The bit-flip mutation operator introduces variability by flipping the bits of a solution's binary representation with a certain mutation probability $p_m$. For a solution $x$ with binary string representation $(x_1, x_2, \ldots, x_l)$, the mutation process is:
\begin{equation}
x_i' = \begin{cases} 
      1 - x_i & \text{with probability } p_m \\
      x_i & \text{with probability } 1 - p_m 
   \end{cases}
\end{equation}
for each bit position $i$ in $\{1, 2, \ldots, l\}$, where $1 - x_i$ represents the binary complement or logical negation of the value of $x_i$. This means each bit $x_i$ has a probability $p_m$ of being flipped (from 0 to 1 or from 1 to 0) and a probability $1 - p_m$ of remaining unchanged.

These operators are used one after another to guide the evolutionary process in GAs.
First, a selection mechanism chooses the parent individuals. Next, crossover operators recombine the parents' genome to generate offspring. Finally, mutation introduces further variations in the offspring's genotype. This GA algorithm constitutes also the basis for advanced optimisation metaheuristics, where specialized TOPSIS-based mutation operators are applied following a conventional bit-flip mutation.

\subsection{Socio-cognitive mutation inspired by TOPSIS}

Socio-cognitive evolutionary algorithm is an optimisation metaheuristic inspired by both social and cognitive aspects of forming a population \cite{ES_SC_article}. To include this additional step, a classical genetic algorithm is expanded by modifying the process of mutation. Instead of one operator present in the classical approach, socio-cognitive algorithm adds more operators allowing individuals to be shaped accordingly to the state of their population. In this article, we adopt the following variants of the TOPSIS-inspired variation operators. These operators modify the mutation method based on the characteristics of the best or worst individuals in the population.

\paragraph{Follow Best}
This operator encourages offspring to adopt traits from the top-performing individuals in the population. For a given offspring solution $x$, the operator randomly selects a ``teacher'' $x^*$ from the set of the top $K$ individuals $\{ x_1, x_2, \ldots, x_K \}$ with the highest fitness values. Each gene $x_i$ in the offspring $x$ is then probabilistically replaced with the corresponding gene $x^*_i$ from the teacher with a given mutation probability $p_m$:
\begin{equation}
x_i = \begin{cases} 
x^*_i & \text{with probability } p_m \\
x_i & \text{with probability } 1 - p_m. 
\end{cases}
\end{equation}

\paragraph{Follow Best Distinct}
This operator refines the Follow Best approach by focusing on genes with high variability among the top $K$ individuals, aiming to standardize these genes in the offspring. For each gene position $i$, the standard deviation $\sigma_i$ across the top individuals is computed:
\begin{equation}
\sigma_i = \sqrt{\frac{1}{K} \sum_{k=1}^K \left( x_{ik} - \bar{x}_i \right)^2},
\end{equation}
where $x_{ik}$ is the value of gene $i$ in the $k$-th individual and $\bar{x}_i$ is the mean value of gene $i$ among the top $K$ solutions. Next, a probability distribution over gene positions is derived using a softmax function:
\begin{equation}
\pi_i = \frac{e^{\sigma_i}}{\sum_{j=1}^l e^{\sigma_j}},
\end{equation}
where $l$ is the total number of genes. A subset of gene positions of size \mbox{$l \times p_m$} is selected based on these probabilities, and for each chosen position $i$, the offspring's gene $x_i$ is replaced with the corresponding gene $x_i^*$ from a randomly selected individual from the top $K$ solutions in the population.

\paragraph{Repel Worst Gravity}
This operator aims to distance offspring from the worst-performing individuals in the population. For a given offspring solution $x$, a ``repeller'' $x^*$ is randomly chosen from the set of the worst $K$ individuals
$\{ x_{n-K+1},\allowbreak \; x_{n-K+2},\allowbreak \ldots, \allowbreak x_n \}$
with the lowest fitness values. Each gene $x_i$ in the offspring is then probabilistically set to the complement of the corresponding gene $x^*_i$ from the repeller:
\begin{equation}\label{eq:repelling}
x_i = \begin{cases} 
1 - x^*_i & \text{with probability } p_m \\
x_i & \text{with probability } 1 - p_m.
\end{cases}
\end{equation}

\paragraph{Repel Worst Gravity Multistep}
This operator extends the Repel Worst Gravity approach by sequentially applying the repulsion effect from multiple worst-performing individuals. For each offspring solution $x$, the operator iterates over the set of the worst $K$ individuals. For each ``repeller'' $x^*$ in this set, each gene $x_i$ in the offspring is probabilistically set to the complement of the corresponding gene $x^*_i$ as described in \eqref{eq:repelling}, allowing the offspring to incrementally diverge from the traits of multiple low-performing individuals.

These operators introduce directed variations in the population, leveraging information about the best and worst individuals to enhance the search process in Genetic Algorithms. In this article, we present the following variants of the algorithm (with abbreviations in parentheses). Each TOPSIS-inspired socio-cognitive mutation operator is applied after standard bit-flip mutation, between lines 6 and 7 in the structure presented in algorithm \ref{algo:ga} above. In case of combined algorithms, the operators were applied sequentially, one after another, in a provided order.
\begin{enumerate}
    \item Follow Best (FB)

    Additional Follow Best mutation is performed.


    
    
    \item Follow Best Distinct (FBD)

    Additional Follow Best Distinct mutation is performed.


    

    \item Repel Worst Gravity (RW)

    Additional Follow Repel Worst Gravity mutation is performed.


    
    \item Repel Worst Gravity Multistep (RWM)

    Additional Repel Worst Gravity Multistep mutation is performed.


    
    \item Follow Best combined with Repel Worst Gravity (FB + RW)

    Follow Best mutation is performed followed by Repel Worst Gravity mutation.
    
    \item Follow Best combined with Repel Worst Gravity Multistep (FB + RWM)

    Follow Best mutation is performed followed by Repel Worst Gravity Multistep mutation.
    
    \item Follow Best Distinct combined with Repel Worst Gravity (FBD + RW)

    Follow Best Distinct mutation is performed followed by Repel Worst Gravity mutation.
    
    \item Follow Best Distinct combined with Repel Worst Gravity Multistep (FBD + RWM)

    Follow Best Distinct mutation is performed followed by Repel Worst Gravity Multistep mutation.
\end{enumerate}

\section{Experiments}

The socio-cognitive  evolutionary algorithms were tested as strategies for solving the low autocorrelation binary sequence (LABS) problem. It is a combinatorial optimization problem that involves finding a binary sequence that has low autocorrelation properties. The goal is to minimize the autocorrelation of sequence.

Formally, a binary sequence in LABS problem of size $L$  is such a sequence $S$ that satisfies $S \in \{-1, 1\}^L$. The aperiodic autocorrelation function of a sequence $S$ with distance $k$ is represented by the equation 
\begin{equation}
C_k(S) = \sum_{i=1}^{L-k} s_i \cdot s_{i+k},
\end{equation}
where $s_i$ is an $i$-th element of a sequence $S$.
The energy function of a sequence $S$ is defined as

\begin{equation}
E(S) = \sum_{k=1}^{L-1} C_k^2(S).
\end{equation} 
The LABS problem of specified size consists of finding a sequence of this size with minimum value of the energy function. \cite{LABS_article}


All tests were conducted against the LABS problem of size 50. This size provides a challenging yet computationally feasible search space to assess the effectiveness of the proposed mutation operators. The operators' parameters were chosen based on preliminary experiments in a way that allowed us to maintain a balance between computational efficiency and solution diversity. The initial population was 20 and the offspring population was 10. A base mutation was a simple bit flip mutation and the simplex crossover was used for crossovers between parents, both with probability of 0.5. The best solution selection algorithm was chosen as a selection algorithm. The termination criterion was reaching 10000 evaluations.

Two versions of each socio-cognitive mechanism were tested: a one where a probability of the mutation for each gene is 0.5 and a one where only one gene is mutated.

\begin{figure}[ht]
    \centering
    \makebox[\textwidth]{%
        \begin{subfigure}[b]{1.0\textwidth}
            \centering
            \includegraphics[width=\textwidth, height=0.50\textheight, keepaspectratio]{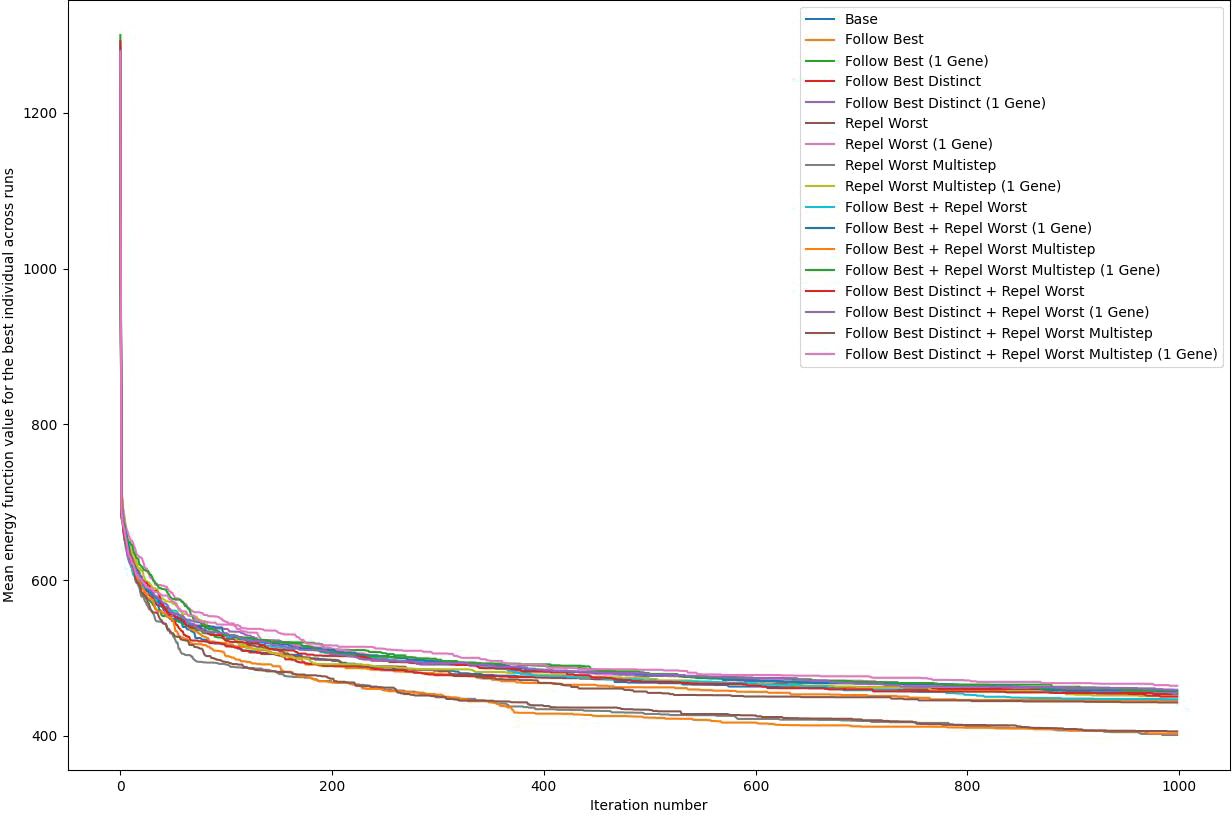}
            \subcaption{Mean energy function values for the best individual in each iteration across runs for each algorithm.}
            \label{fig:mean_all}
        \end{subfigure}
    }
    
    \makebox[\textwidth]{%
        \begin{subfigure}[b]{1.0\textwidth}
            \centering
            \includegraphics[width=\textwidth, height=0.50\textheight, keepaspectratio]{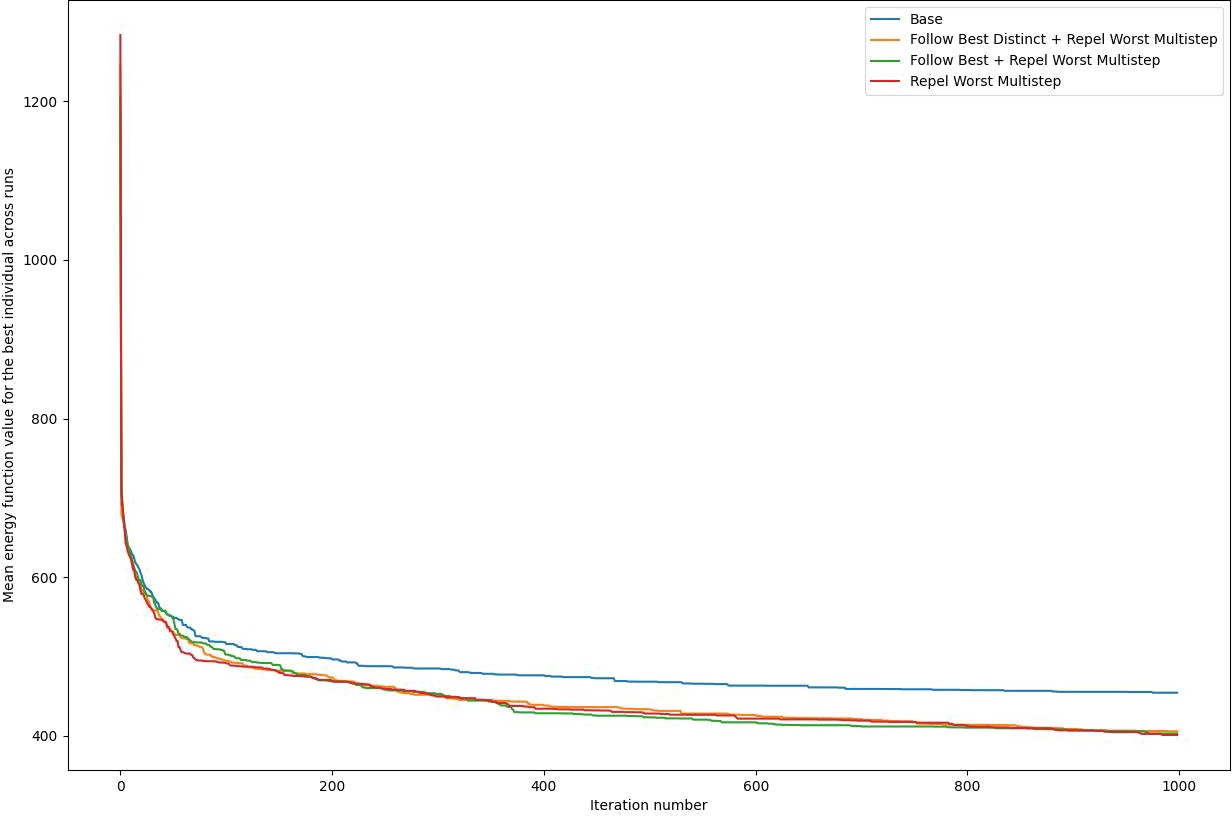}
            \subcaption{Mean energy function values for the best individual in each iteration across runs for algorithms with significantly different results.}
            \label{fig:mean_relevant}
        \end{subfigure}
    }

    \caption{Comparison of mean energy function values for different algorithm variants. (a) Includes all algorithms. (b) Focuses on significantly different results.}
    \label{fig:combined}
\end{figure}

Each algorithm, including a base one with no socio-cognitive mechanisms added, was run 50 times. Mean energy function values for the best individual in each iteration across runs are displayed in Figure \ref{fig:mean_all}. Additional numerical data is presented in the tables below.

\begin{table}[ht]
\centering
\renewcommand{\arraystretch}{1.2} 
\setlength{\extrarowheight}{2pt}  

\begin{tabularx}{\linewidth}{|l||>{\centering\arraybackslash}X|>{\centering\arraybackslash}X|>{\centering\arraybackslash}X|>{\centering\arraybackslash}X|>{\centering\arraybackslash}X|>{\centering\arraybackslash}X|>{\centering\arraybackslash}X|>{\centering\arraybackslash}X|>{\centering\arraybackslash}X|}
 \hline
 Algorithm & Base & FB & FBD & RW & RWM & FB + RW & FB + RWM & FBD + RW & FBD + RWM \\
 \hline
 Value & 454.44 & 445.32 & 453.8 & 442.92 & 401.16 & 447.08 & 403.08 & 450.44 & 405.8 \\
 \hline
\end{tabularx}

\caption{Mean energy function value for best individual at the end of a run across runs for each algorithm (variants mutation probability of 0.5 for each gene).}
\end{table}

\begin{table}[ht]
\centering
\renewcommand{\arraystretch}{1.2} 
\setlength{\extrarowheight}{2pt}  

\begin{tabularx}{\linewidth}{|l||>{\centering\arraybackslash}X|>{\centering\arraybackslash}X|>{\centering\arraybackslash}X|>{\centering\arraybackslash}X|>{\centering\arraybackslash}X|>{\centering\arraybackslash}X|>{\centering\arraybackslash}X|>{\centering\arraybackslash}X|>{\centering\arraybackslash}X|}
 \hline
 Algorithm & Base & FB & FBD & RW & RWM & FB + RW & FB + RWM & FBD + RW & FBD + RWM \\
 \hline
 Value & 454.44 & 457.16 & 453.32 & 456.36 & 451.08 & 457.16 & 458.12 & 459.24 & 464.36 \\ 
 \hline
\end{tabularx}

\caption{Mean energy function value for best individual at the end of a run across runs for each algorithm (variants with a mutation of a single gene).}
\end{table}

The values of the energy function for the best individual at the end of each run were tested against the corresponding values for the base algorithm using the Wilcoxon test. Setting the significance level to 0.05 and comparing it with the calculated p-values, it was concluded that the only algorithms that give results that are statistically significantly different from the base one were:
\begin{itemize}
    \item follow best distinct with repel worst multistep,
    \item follow best with repel worst multistep,
    \item repel worst multistep,
\end{itemize}
each with a 0.5 probability of the mutation of each gene.

Mean energy function values for the best individual in each iteration across runs for the aforementioned algorithms are displayed in Figure \ref{fig:mean_relevant}. A standard boxplot with whiskers of the energy function values of the best individual at the end of each run for each relevant algorithm is presented below.

\begin{figure}[ht]
\centering
\includegraphics[width=.8\textwidth]{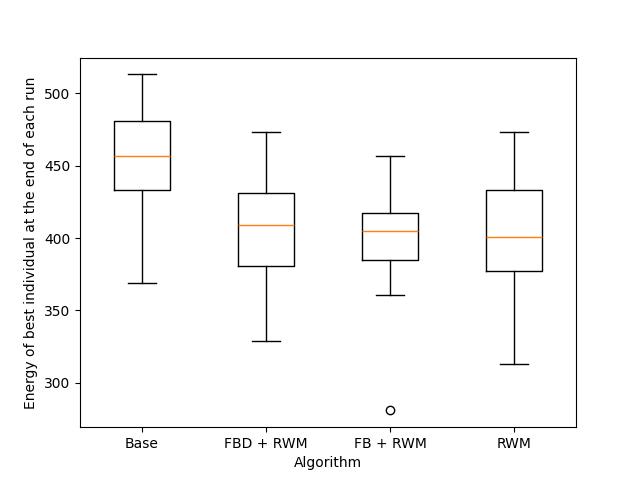}
\caption{A standard boxplot with whiskers of the energy function values of the best individual at the end of each run for each algorithm that obtains significantly different results from the base algorithm.}
\end{figure}

The best (lowest) of the energy values for each algorithm was:
\begin{itemize}
    \item 329 for follow best distinct with repel worst multistep,
    \item 281 for follow best with repel worst multistep,
    \item 313 for repel worst multistep,
\end{itemize}
while for the base algorithm the value was 369.

\section{Conclusions}
This study introduced a socio-cognitive mutation operators based on the TOPSIS method, designed to enhance the search efficiency of evolutionary algorithms applied to the LABS problem. By integrating principles of observational learning, imitation, and adaptive decision-making, the proposed mutation strategies: the ``Repel Worst Multistep'', the ``Follow Best Distinct with Repel Worst Multistep'' and the ``Follow Best with Repel Worst Multistep'' demonstrated improved performance over traditional approaches.

Key findings include:
\begin{itemize}
    \item Follow Best Distinct with Repel Worst Gravity Multistep yielded the best results, significantly reducing autocorrelation values.
    \item Avoiding poor solutions (repulsion mechanisms) contributed to maintaining population diversity, preventing premature convergence.
\end{itemize}
These results reinforce the potential of socio-cognitive metaheuristics in advancing evolutionary computation. Future work should explore the application of these mechanisms to real-world combinatorial optimization problems, such as scheduling and cryptographic sequence generation.

\subsection*{Acknowledgements}

The research presented in this paper has been financially supported by: Polish National Science Center Grant no. 2019/35/O/ST6/00570 ``Socio-cognitive inspirations in classic metaheuristics.'';  Polish Ministry of Science and Higher Education funds assigned to AGH University of Science and Technology. ARTIQ project – Polish National Science Center:DEC-2021/01/2/ST6/00004, Polish National Center for Research and Development, DWP/ARTIQI/426/2023 (AB)


%
%
\bibliographystyle{splncs04}
\bibliography{bibliography}

\end{document}